\documentclass[runningheads, orivec]{llncs}
\usepackage[T1]{fontenc}
\usepackage{graphicx,verbatim}
\usepackage{booktabs}
\usepackage{colortbl}
\usepackage{xcolor}
\usepackage{multirow}
\usepackage{amsmath}
\usepackage{amssymb}
\usepackage{microtype}
\usepackage{pifont}
\usepackage{dsfont}
\newcommand{\cmark}{\ding{51}}

\begin{document}

\title{Distilling Temporal Coherence into 2D Networks for Transrectal Ultrasound Prostate Video Segmentation}
\titlerunning{Distilling Temporal Coherence into 2D Networks}

\author{
  Dong Yeong Kim$^{\ast}$\inst{1,2} \and 
  JunGyu Lee$^{\ast}$\inst{3}$^{\dagger}$ \and 
  Jaewon Choi\inst{2} \and 
  June Young Seo\inst{4} \and \\ 
  Myeongseop Kim\inst{2} \and 
  Jinwook Choi\inst{1} \and 
  Taek Min Kim\inst{4}$^{\ddagger}$ \and 
  Young-Gon Kim\inst{2}$^{\ddagger}$
}


\authorrunning{D.Y. Kim and J. Lee et al.}


\institute{
  $^1$ Interdisciplinary Program in Bioengineering, Seoul National University \quad \\ 
  $^2$ Department of Transdisciplinary Medicine, Seoul National University Hospital \quad \\
  $^3$ Department of Artificial Intelligence, Yonsei University \quad \\
  $^4$ Department of Radiology, Seoul National University Hospital \\
  \email{steven6774@snu.ac.kr, jungyu.lee@yonsei.ac.kr}
}

\maketitle

\let\thefootnote\relax\footnotetext{\mbox{$^{\ast}$ Equal contribution \  $^{\ddagger}$ Corresponding Author \  $^{\dagger}$ Work done prior to joining\inst{3}}}
\begin{abstract}

Real-time video segmentation of the prostate in Transrectal Ultrasound (TRUS) is essential for image-guided interventions. While conventional 2D methods suffer from inter-frame inconsistencies by disregarding temporal context, 3D architectures incur prohibitive latency. To resolve this dilemma, we present a Temporally Consistent Learning Framework that distills temporal coherence into a 2D network during training, preserving single-frame inference efficiency. Our design is driven by a key clinical observation: the prostate exhibits geometric stability, whereas the surrounding acoustic environment fluctuates due to physiological motion and transducer pressure. Because conventional temporal constraints propagate erroneous gradients from these unstable regions, we introduce a Confidence-Weighted Temporal Consistency objective derived from optical flow warping residuals, selectively attenuating contributions from unreliable regions. Complementing this pixel-wise constraint, a Dual-scale Prototype Alignment Module enforces semantic coherence through contrastive optimization of local boundary and global semantic features. Furthermore, to eliminate the need for dense per-frame video annotations, we employ geometric equivariance-based pseudo-labeling with knowledge distillation from a pretrained teacher. Extensive experiments on SUN-SEG and our newly introduced TRUS-V benchmark (2,679 frames) demonstrate state-of-the-art accuracy and temporal consistency at real-time speed. Code and dataset are available at https://github.com/DYDevelop/DTC-TRUS.

\end{abstract}

\section{Introduction}

Prostate cancer remains a leading cause of male malignancy worldwide~\cite{siegel2023cancer}, necessitating reliable imaging for early diagnosis and intervention. Transrectal Ultrasound (TRUS) has established itself as the standard modality for procedures such as biopsy and brachytherapy, owing to its real-time capability, cost-effectiveness, and radiation-free nature~\cite{lei2019ultrasound}. In clinical workflows, accurate segmentation of the prostate boundary from TRUS videos is a prerequisite for precise volume estimation and treatment planning. However, relying on manual delineation is labor-intensive and prone to inter-observer variability. This challenge is further exacerbated by the inherent artifacts of TRUS, such as low contrast, severe speckle noise, and ambiguous boundaries, which complicate the development of robust real-time automatic segmentation systems~\cite{wang2024review}.

\begin{figure}[t]
    \centering
    \includegraphics[trim={0mm 35mm 0mm 30mm},clip,width=1\linewidth]{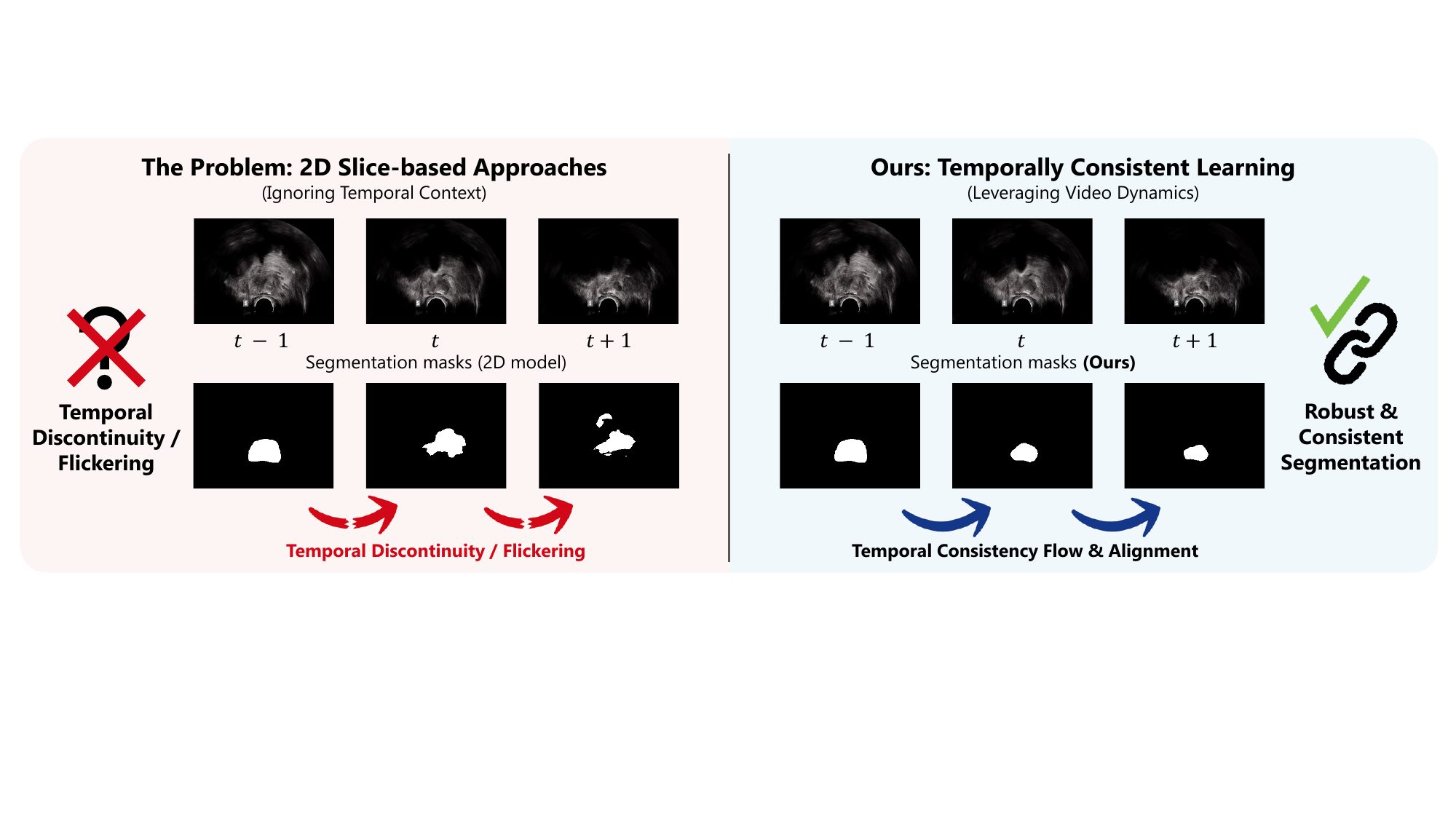}
    \vspace{-14mm}
    \caption{\textbf{Impact of the proposed consistency learning.} Illustration of the limitations of 2D slice-based methods (left) versus our temporally consistent approach (right). Our method significantly reduces flickering artifacts and improves boundary robustness in challenging prostate ultrasound videos.}
    \vspace{-4mm}
    \label{fig:1}
    
\end{figure}

Recent advances in deep learning have achieved remarkable success in medical image segmentation~\cite{ronneberger2015u}. However, most existing approaches for prostate segmentation treat video data as a set of independent 2D static images~\cite{karimi2019accurate,lei2019multidirectional,ghavami2019integration}. While computationally efficient, these 2D slice-based models inherently neglect temporal context, leading to unstable predictions across adjacent frames, commonly manifesting as ``flickering'' artifacts that are unacceptable for clinical applications requiring smooth tracking (Fig.~\ref{fig:1}). To capture spatiotemporal features, 3D CNNs and Recurrent Neural Networks have been employed~\cite{milletari2016v,wang2019deep,yang2017fine,orlando2020automatic}. However, these methods incur high computational costs, rendering them impractical for intra-operative procedures where low-latency feedback is strictly mandated~\cite{wang2024review}. To balance accuracy and efficiency, a third line of work augments a 2D feature extractor with a dedicated temporal-interaction module~\cite{hu2024sali,xu2024lgrnet}. Nevertheless, such designs still introduce auxiliary temporal modules at inference and rely on temporally annotated video for supervision, leaving the trade-off between runtime and annotation cost unresolved.

In this work, we propose a \textbf{Temporally Consistent Learning Framework} that distills temporal coherence into a standard 2D network during training, preserving single-frame inference efficiency. A key observation motivating our design is that the prostate exhibits geometric stability, whereas the surrounding acoustic environment fluctuates substantially due to physiological motion and transducer pressure. Consequently, naive temporal constraints propagate erroneous gradients from unreliable background regions, degrading rather than improving consistency.

Specifically, we introduce two synergistic mechanisms. First, a \textbf{Confidence-Weighted Temporal Consistency} objective incorporates a weighting scheme derived from optical flow warping residuals. This mechanism selectively attenuates contributions from dynamically unstable regions, concentrating supervisory signals on the anatomically consistent prostate structure. Second, a \textbf{Dual-scale Prototype Alignment Module} enforces semantic coherence in the latent space through contrastive optimization of both local boundary features and global semantic representations across temporal neighbors, ensuring robustness against speckle noise and intensity variations.

To alleviate the burden of dense per-frame video annotation, we employ a self-supervised strategy leveraging geometric equivariance for pseudo-label generation. Additionally, a Student-Teacher architecture with knowledge distillation mitigates catastrophic forgetting of spatial priors learned from labeled static images. Furthermore, we introduce \textbf{TRUS-V}, a large-scale video benchmark comprising 2,679 annotated frames across axial and sagittal views from 10 patients, thereby addressing the scarcity of publicly available video datasets in this domain.

\noindent Our main contributions are summarized as follows:
\begin{itemize}
    \item We propose a Temporally Consistent Learning Framework that distills temporal coherence into a 2D segmentation model during training, eliminating the need for heavy 3D computations or temporal-interaction modules at inference while achieving temporally stable predictions.
    \item We introduce Confidence-Weighted Temporal Consistency and Dual-scale Prototype Alignment, which synergistically suppress flickering artifacts and enhance boundary robustness by focusing on stable anatomical structures.
    \item We release TRUS-V, a large-scale benchmark comprising 2,679 densely annotated frames (Axial/Sagittal) from 10 patients. Extensive experiments on TRUS-V and SUN-SEG demonstrate state-of-the-art performance in both accuracy and temporal coherence.
\end{itemize}
\section{Datasets}

\noindent\textbf{Newly Collected TRUS-V Benchmark.}
To address the lack of multi-view video datasets for prostate segmentation, we introduce TRUS-V, an in-house benchmark comprising 2,679 frames collected from 10 patients. Unlike existing image-based datasets, TRUS-V captures temporal dynamics through 20 continuous video sequences, consisting of paired Axial and Sagittal views for each patient. To ensure rigorous evaluation, we partitioned the dataset at the patient level, yielding 2,400 training and 279 testing frames. 
Given the prohibitive cost of dense manual annotation for video data, we established the ground truth using a semi-automated annotation strategy. First, we utilized a separate static dataset containing 2,140 Axial and 2,260 Sagittal images to train a 5-fold ensemble U-Net, which achieved a high mean Dice Similarity Coefficient (DSC) of 0.95 (Axial) and 0.93 (Sagittal). This model generated initial candidate masks for the TRUS-V video sequences. Subsequently, experienced radiologists visually inspected and manually refined these masks frame-by-frame to correct inaccuracies. This human-in-the-loop process ensures the high reliability of our reference standard.

\noindent\textbf{Public Dataset: SUN-SEG.}
To assess generalization beyond the ultrasound modality, we utilized SUN-SEG~\cite{ji2022video}, a large-scale Video Polyp Segmentation benchmark (158,690 frames), characterized by diverse challenges such as fast motion and occlusion. We followed the dataset's standard training and evaluation protocols to validate that our Temporally Consistent Learning Framework is effective for general video segmentation tasks requiring temporal coherence.
\section{Methods}

\begin{figure*}[t]
    \centering
    \includegraphics[trim={25mm 25mm 25mm 25mm},clip,width=1\linewidth]{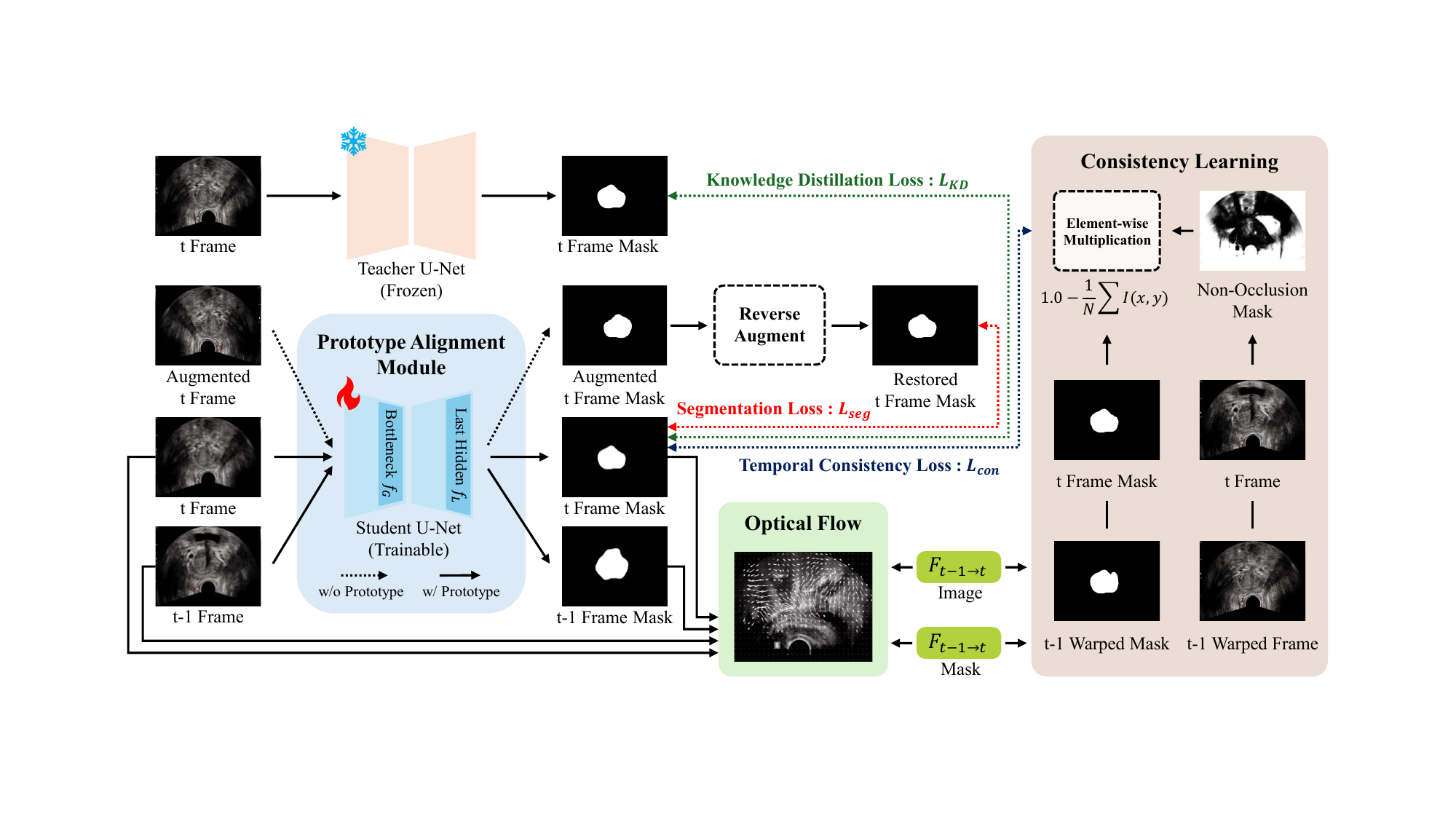}
    \vspace{-7mm}
    \caption{\textbf{Overview of the proposed framework.} The network distills temporal coherence via Confidence-Weighted Temporal Consistency and Dual-scale Prototype Alignment. We employ a self-supervised Student-Teacher strategy to utilize unlabeled videos via pseudo-labeling and knowledge distillation. Crucially, only the efficient 2D Student is required during inference, ensuring real-time performance.}
    \label{fig:3}
    \vspace{-3mm}
\end{figure*}

\subsection{Overview of the Framework}
To achieve real-time video segmentation without the computational overhead of 3D networks, we propose a Temporally Consistent Learning Framework (see Fig.~\ref{fig:3}) that empowers a standard 2D backbone to capture video dynamics. Given an unlabeled video sequence $\mathcal{V} = \{I_1, \dots, I_T\}$, our objective is to train a Student model $\mathcal{S}(\cdot; \theta_S)$ that generates accurate and consistent masks $P_t = \mathcal{S}(I_t)$ without frame-wise annotations. To achieve this, we employ a Student-Teacher architecture enforcing coherence at two synergistic levels: (1) pixel-level geometric alignment via optical flow to explicitly model tissue deformation, and (2) feature-level semantic alignment via prototypes to robustly suppress speckle noise. This bi-level strategy ensures robust temporal representation learning while maintaining high inference efficiency.

\subsection{Self-Supervised Equivariance and Distillation}
\label{sec:self_sup}
To circumvent the prohibitive cost of dense per-frame video annotation, we adopt
a self-supervised strategy anchored by geometric equivariance~\cite{bortsova2019semi} and knowledge distillation~\cite{hinton2015distilling}. Notably, the video adaptation stage requires
no frame-wise video labels; spatial priors are instead inherited from a Teacher
pre-trained solely on the separate static image dataset. First, relying on the hypothesis that segmentation should be equivariant to geometric transformations, we generate pseudo-labels $\hat{Y}_t$ by applying the inverse transformation to predictions of a flipped input $\mathcal{A}(I_t)$. The student minimizes $\mathcal{L}_{seg} = \mathcal{H}(P_t, \text{stop\_grad}(\hat{Y}_t))$, where $\mathcal{H}$ denotes the BCE-Dice loss and the stop-gradient prevents mode collapse~\cite{chen2021exploring}. Simultaneously, to mitigate catastrophic forgetting of spatial representations, we distill knowledge from a frozen static Teacher (sharing the identical architecture with the Student) using the $\ell_2$-based objective $\mathcal{L}_{KD} = || \sigma(Z_S/\tau) - \sigma(Z_T/\tau) ||^2_2$, where $\sigma$ represents the sigmoid function, $\tau$ is the temperature parameter, and $Z_S$ and $Z_T$ denote the student and teacher logits, respectively. This effectively regularizes training, ensuring the model retains robust anatomical priors while adapting to video dynamics.

\begin{figure*}[t]
    \centering
    \includegraphics[trim={7mm 33mm 0mm 26mm},clip,width=1\linewidth]{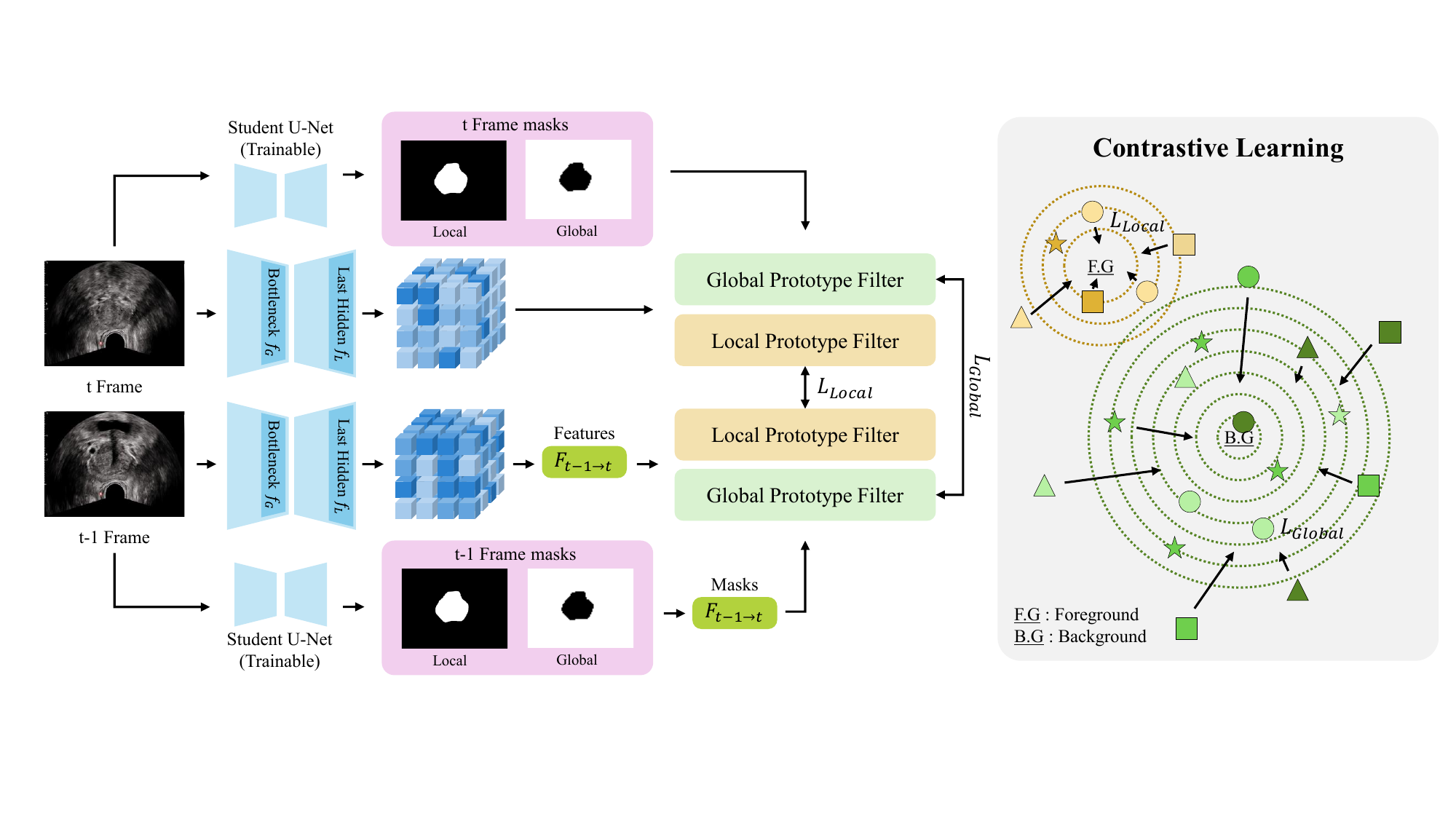}
    \vspace{-6mm}
    \caption{\textbf{Dual-scale Prototype Alignment Module.} We extract global and local prototypes to enforce semantic consistency. A temporal alignment objective (Right) optimizes intra-class compactness by matching features across adjacent frames, ensuring robust target boundaries and scene stability against video artifacts.}
    \label{fig:4} 
    \vspace{-3mm}
\end{figure*}

\subsection{Confidence-Weighted Temporal Consistency}
\label{sec:flow}
To explicitly model the temporal continuity of the prostate anatomy, we introduce a motion-aware consistency loss. We estimate the optical flow field $\mathcal{F}_{t-1 \to t}$ between adjacent frames $I_{t-1}$ and $I_t$ using a dense flow estimation algorithm (specifically, Farneback~\cite{farneback2003two}).
The predicted mask of the previous frame $P_{t-1}$ is warped to the current coordinate system using a warping operation $\mathcal{W}$~\cite{jaderberg2015spatial}:
\begin{equation}
    \tilde{P}_{t-1 \to t} = \mathcal{W}(P_{t-1}, \mathcal{F}_{t-1 \to t})
\end{equation}
We enforce the warped mask $\tilde{P}_{t-1 \to t}$ to be consistent with the current prediction $P_t$. To handle occlusions or motion estimation errors, we compute a non-occlusion confidence map $M_{noc} = \exp(-|I_t - \mathcal{W}(I_{t-1}, \mathcal{F}_{t-1 \to t})|)$.
Unlike standard weighted $\ell_1$ losses, we formulate the objective to maximize the structural similarity in reliable regions~\cite{lai2018learning}. The temporal consistency loss is defined as:
\begin{equation}
    \mathcal{L}_{con} = 1 - \frac{1}{N} \sum_{p} M_{noc}(p) \cdot \left( 1 - | P_t(p) - \tilde{P}_{t-1 \to t}(p) | \right)
\end{equation}
where $N$ is the total number of pixels and $p$ denotes the pixel index. This formulation minimizes the loss when the predictions are consistent ($| P_t - \tilde{P}_{t-1 \to t} | \approx 0$) in regions with high confidence ($M_{noc} \approx 1$), while effectively ignoring gradients in occluded or unstable regions where $M_{noc} \approx 0$.

\subsection{Dual-scale Prototype Alignment Module}
\label{sec:proto}
Pixel-level consistency alone may be insufficient under severe speckle noise or acoustic shadows. Therefore, we propose a Dual-scale Prototype Alignment Module to enforce semantic consistency across different feature granularities. We extract prototypes from both Global (Bottleneck) and Local (Decoder) features of the Student network (see Fig.~\ref{fig:4}).

\noindent\textbf{Prototype Extraction.} Let $f_t \in \{f^{loc}_t, f^{glob}_t\}$ be the local or global features. A prototype $v_{c,t}$ for class $c \in \{c_f, c_b\}$ is computed via masked average pooling~\cite{snell2017prototypical}:
\begin{equation}
    v_{c, t} = \frac{\sum_{x,y} \mathds{1}_c(x,y) \cdot f_t(x,y)}{\sum_{x,y} \mathds{1}_c(x,y) + \epsilon}
\end{equation}
where $\mathds{1}_c(\cdot)$ is a binary indicator isolating the foreground ($c_f$) or background ($c_b$) via an adaptive threshold.

\noindent\textbf{Dual-scale Contrastive Alignment.}
To explicitly model multi-scale dynamics, we align $c_f$ prototypes on local features (preserving boundaries) and $c_b$ prototypes on global features (stabilizing scenes). We warp previous local features $f^{loc}_{t-1}$ to $t$ ($\tilde{v}^{loc}$) and directly compare spatially sparse global ones, whose high-level semantics are naturally motion-invariant—to maximize intra-class cosine similarity~\cite{wang2019panet}. To handle scale variations dynamically, we cross-weight using background ($w_{c_b}$) and foreground ($w_{c_f}$) mask fractions:
\begin{equation}
    \mathcal{L}_{proto} = w_{c_b} \big( 1 - \cos(\tilde{v}^{loc}_{c_f, t-1}, v^{loc}_{c_f, t}) \big) + w_{c_f} \big( 1 - \cos(v^{glob}_{c_b, t-1}, v^{glob}_{c_b, t}) \big)
\end{equation}
This elegantly balances alignment; smaller targets (low $w_{c_f}$) receive higher local boundary weights ($w_{c_b}$), enforcing robust intra-class compactness.

\subsection{Total Objective Function}
The final objective function is a weighted sum of the aforementioned losses:
\begin{equation}
    \mathcal{L}_{total} = \lambda_{seg}\mathcal{L}_{seg} + \lambda_{KD}\mathcal{L}_{KD} + \lambda_{con}\mathcal{L}_{con} + \lambda_{proto}\mathcal{L}_{proto}
\end{equation}
where $\lambda_{seg}$, $\lambda_{KD}$, $\lambda_{con}$, and $\lambda_{proto}$
are balancing coefficients. During inference, only the Student 2D network is employed, maintaining the computational efficiency of standard 2D models without calculating optical flow or teacher inference.

\section{Experiments}
\subsection{Implementation Details}
Our framework is backbone-agnostic; we instantiate it with U-Net++~\cite{zhou2019unet++}
on TRUS-V and ACSNet~\cite{zhang2020adaptive} on SUN-SEG, all trained on a single
NVIDIA A100 40GB. The static Teacher is pre-trained on a separate static image dataset and frozen, after which the Student is adapted on unlabeled videos with AdamW
(learning rate $1\times10^{-6}$, batch size $8$) for one epoch. We set
$\lambda_{seg}{=}\lambda_{KD}{=}\lambda_{proto}{=}1.0$, and
$\lambda_{con}{=}0.25$.
\vspace{-3mm}
\begin{table*}[ht!]
\caption{Quantitative comparison on SUN-SEG dataset. The best and second-best results are highlighted in \textbf{bold} and \underline{underlined}, respectively.}
\vspace{-3mm}
\label{tab:comparison}
\centering
\setlength{\aboverulesep}{0pt}     
\setlength{\belowrulesep}{0pt}     

\resizebox{\textwidth}{!}{%
\begin{tabular}{cc|cccccc|cccccc}
\toprule
\multicolumn{2}{c|}{\multirow{2}{*}{Method}} & \multicolumn{6}{c|}{SUN-SEG-Easy (Unseen)} & \multicolumn{6}{c}{SUN-SEG-Hard (Unseen)} \\ \cline{3-14} 
\multicolumn{2}{c|}{} & $S_{\alpha} \uparrow$ & $E_{\phi}^{mn} \uparrow$ & $F_{\beta}^w \uparrow$ & $F_{\beta}^{mn} \uparrow$ & Dice $\uparrow$ & Sen $\uparrow$ & $S_{\alpha} \uparrow$ & $E_{\phi}^{mn} \uparrow$ & $F_{\beta}^w \uparrow$ & $F_{\beta}^{mn} \uparrow$ & Dice $\uparrow$ & Sen $\uparrow$ \\ \midrule
\multicolumn{1}{c|}{\multirow{9}{*}{\rotatebox{90}{Image Models}}} & U-Net~\cite{ronneberger2015u} & 0.669 & 0.677 & 0.459 & 0.528 & 0.530 & 0.420 & 0.670 & 0.679 & 0.457 & 0.527 & 0.542 & 0.429 \\
\multicolumn{1}{c|}{} & U-Net$^{++}$~\cite{zhou2019unet++} & 0.684 & 0.687 & 0.491 & 0.553 & 0.559 & 0.457 & 0.685 & 0.697 & 0.480 & 0.544 & 0.554 & 0.467 \\
\multicolumn{1}{c|}{} & ACSNet~\cite{zhang2020adaptive} & 0.782 & 0.779 & 0.642 & 0.688 & 0.713 & 0.601 & 0.793 & 0.787 & 0.636 & 0.684 & 0.708 & 0.618 \\
\multicolumn{1}{c|}{} & PraNet~\cite{fan2020pranet} & 0.733 & 0.753 & 0.572 & 0.632 & 0.621 & 0.524 & 0.717 & 0.735 & 0.544 & 0.607 & 0.598 & 0.512 \\
\multicolumn{1}{c|}{} & ColonSegNet~\cite{jha2021real} & 0.776 & 0.768 & 0.621 & 0.666 & 0.683 & 0.594 & 0.789 & 0.770 & 0.636 & 0.662 & 0.685 & 0.599 \\
\multicolumn{1}{c|}{} & MedNeXt~\cite{roy2023mednext} & 0.703 & 0.719 & 0.553 & 0.598 & 0.591 & 0.506 & 0.699 & 0.701 & 0.522 & 0.575 & 0.611 & 0.508 \\
\multicolumn{1}{c|}{} & MSRF-Net~\cite{srivastava2021msrf} & 0.794 & 0.780 & 0.652 & 0.693 & 0.701 & 0.600 & 0.774 & 0.762 & 0.637 & 0.658 & 0.701 & 0.605 \\
\multicolumn{1}{c|}{} & TransNetR~\cite{jha2024transnetr} & 0.768 & 0.768 & 0.648 & 0.694 & 0.642 & 0.592 & 0.766 & 0.772 & 0.629 & 0.673 & 0.628 & 0.591 \\
\multicolumn{1}{c|}{} & \cellcolor[gray]{0.9}\textbf{Ours} & \cellcolor[gray]{0.9}\underline{0.816} & \cellcolor[gray]{0.9}\textbf{0.882} & \cellcolor[gray]{0.9}\textbf{0.738} & \cellcolor[gray]{0.9}\textbf{0.784} & \cellcolor[gray]{0.9}\underline{0.746} & \cellcolor[gray]{0.9}\underline{0.719} & \cellcolor[gray]{0.9}\textbf{0.816} & \cellcolor[gray]{0.9}\textbf{0.878} & \cellcolor[gray]{0.9}\textbf{0.719} & \cellcolor[gray]{0.9}\textbf{0.758} & \cellcolor[gray]{0.9}\textbf{0.737} & \cellcolor[gray]{0.9}\textbf{0.741} \\ \midrule
\multicolumn{1}{c|}{\multirow{3}{*}{\rotatebox{90}{\shortstack{Video\\Models}}}} & SSTAN~\cite{zhao2022semi} & 0.805 & 0.838 & 0.691 & 0.745 & 0.726 & 0.662 & 0.801 & 0.733 & \underline{0.682} & 0.734 & 0.718 & \underline{0.676} \\
\multicolumn{1}{c|}{} & FLA-Net~\cite{lin2023shifting} & 0.722 & 0.697 & 0.547 & 0.597 & 0.636 & 0.506 & 0.721 & 0.701 & 0.540 & 0.592 & 0.628 & 0.522 \\
\multicolumn{1}{c|}{} & DALA~\cite{zhao2025efficient} & \textbf{0.837} & \underline{0.854} & \underline{0.722} & \underline{0.772} & \textbf{0.768} & \textbf{0.721} & \underline{0.808} & \underline{0.840} & 0.676 & \underline{0.735} & \underline{0.733} & 0.669 \\ \bottomrule
\end{tabular}%
}
\end{table*}

\begin{table*}[t]
    \centering
    \setlength{\aboverulesep}{0pt}
    \setlength{\belowrulesep}{0pt}
    
    \begin{minipage}[t]{0.492\textwidth}
        \caption{Quantitative comparison on the TRUS-V dataset.}
        \label{tab:trus_v}
        \centering
        \setlength{\tabcolsep}{3pt}
        \resizebox{\textwidth}{!}{%
        \begin{tabular}{cc|cccccc}
            \toprule
            \multicolumn{2}{c|}{\multirow{2}{*}{Method}} & \multicolumn{6}{c}{TRUS-V} \\ \cline{3-8} 
            \multicolumn{2}{c|}{} & $S_{\alpha} \uparrow$ & $E_{\phi}^{mn} \uparrow$ & $F_{\beta}^w \uparrow$ & $F_{\beta}^{mn} \uparrow$ & Dice $\uparrow$ & Sen $\uparrow$ \\ \midrule
            \multicolumn{1}{c|}{\multirow{9}{*}{\rotatebox{90}{Image Models}}} & U-Net & 0.964 & 0.985 & 0.808 & 0.814 & 0.817 & 0.829 \\
            \multicolumn{1}{c|}{} & U-Net$^{++}$ & \textbf{0.967} & \textbf{0.988} & \underline{0.821} & 0.820 & 0.820 & 0.827 \\
            \multicolumn{1}{c|}{} & ACSNet & \underline{0.965} & 0.983 & 0.802 & \underline{0.827} & 0.820 & 0.813 \\
            \multicolumn{1}{c|}{} & PraNet & 0.932 & 0.934 & 0.751 & 0.753 & 0.749 & 0.746 \\
            \multicolumn{1}{c|}{} & ColonSegNet & 0.959 & 0.981 & 0.806 & 0.810 & 0.808 & 0.812 \\
            \multicolumn{1}{c|}{} & MedNeXt & 0.886 & 0.924 & 0.771 & 0.799 & 0.795 & 0.832 \\
            \multicolumn{1}{c|}{} & MSRF-Net & \underline{0.965} & \underline{0.987} & \underline{0.821} & 0.816 & \underline{0.821} & \underline{0.838} \\
            \multicolumn{1}{c|}{} & TransNetR & 0.964 & 0.982 & 0.815 & 0.817 & 0.816 & 0.821 \\
            \multicolumn{1}{c|}{} & \cellcolor[gray]{0.9}\textbf{Ours} & \cellcolor[gray]{0.9}\textbf{0.967} & \cellcolor[gray]{0.9}\textbf{0.988} & \cellcolor[gray]{0.9}\textbf{0.830} & \cellcolor[gray]{0.9}\textbf{0.828} & \cellcolor[gray]{0.9}\textbf{0.829} & \cellcolor[gray]{0.9}\textbf{0.839} \\ \midrule
            \multicolumn{1}{c|}{\multirow{3}{*}{\rotatebox{90}{\shortstack{Video\\Models}}}} & SSTAN & 0.963 & 0.980 & 0.816 & 0.814 & 0.819 & 0.837 \\
            \multicolumn{1}{c|}{} & FLA-Net & 0.963 & 0.978 & 0.789 & 0.817 & 0.811 & 0.809 \\
            \multicolumn{1}{c|}{} & DALA & 0.908 & 0.931 & 0.539 & 0.738 & 0.729 & 0.746 \\ \bottomrule
        \end{tabular}%
        }
    \end{minipage}
    \hfill
    \begin{minipage}[t]{0.488\textwidth}
        \caption{Ablation study on the SUN-SEG-Easy (Unseen) dataset to investigate the effectiveness of each proposed component. Proto. refers to the prototype, and $\triangle$ indicates partial (single-scale) usage.}
        \label{tab:ablation}
        \centering
        \setlength{\tabcolsep}{4pt}
        \resizebox{\textwidth}{!}{%
        \begin{tabular}{lcccc|cc}
            \toprule
            \multirow{2}{*}{\textbf{Method}} & \multirow{2}{*}{\textbf{$L_{seg}$}} & \multirow{2}{*}{\textbf{$L_{KD}$}} & \multirow{2}{*}{\textbf{$L_{proto}$}} & \multirow{2}{*}{\textbf{$L_{con}$}} & \multirow{2}{*}{$S_{\alpha} \uparrow$} & \multirow{2}{*}{\textbf{Dice $\uparrow$}} \\
             & & & & & & \\ \midrule
            Baseline & \cmark & - & - & - & 0.274 & 0.171 \\
            w/o Temporal & \cmark & \cmark & - & - & 0.793 & 0.701 \\
            \midrule
            w/o Consistency & \cmark & \cmark & \cmark & - & \underline{0.813} & \underline{0.735} \\ 
            w/o Proto. & \cmark & \cmark & - & \cmark & 0.810 & 0.722 \\
            w/o Global Proto. & \cmark & \cmark & $\triangle$ & \cmark & 0.808 & 0.721 \\
            w/o Local Proto. & \cmark & \cmark & $\triangle$ & \cmark & 0.808 & 0.722 \\
            \midrule
            \rowcolor[gray]{0.9} \textbf{Proposed} & \cmark & \cmark & \cmark & \cmark & \textbf{0.816} & \textbf{0.746} \\ \bottomrule
        \end{tabular}%
        }
    \end{minipage}
\end{table*}
\vspace{-6mm}
\subsection{Comparisons with State-of-the-art Methods}

\noindent\textbf{Performance on Public Benchmark.} 
We evaluate our framework against SOTA methods on the SUN-SEG dataset~\cite{ji2022video}. Following the standard protocol, we report Structure ($S_{\alpha}$), Enhanced-alignment ($E_{\phi}^{mn}$), weighted F-measures ($F_{\beta}^{w}, F_{\beta}^{mn}$), Dice, and Sensitivity (Sen). For fair comparison, all 2D models follow standard frame-by-frame inference. As Table~\ref{tab:comparison} shows, our method not only significantly outperforms 2D baselines but also rivals or surpasses heavy video-based models. This confirms our strategy effectively transfers video dynamics to a lightweight 2D backbone, achieving real-time inference (89.95 FPS via ACSNet~\cite{zhang2020adaptive}) without heavy 3D computations.

\begin{figure}[t]
    \centering
    \includegraphics[trim={5mm 35mm 5mm 32mm},clip,width=1\linewidth]{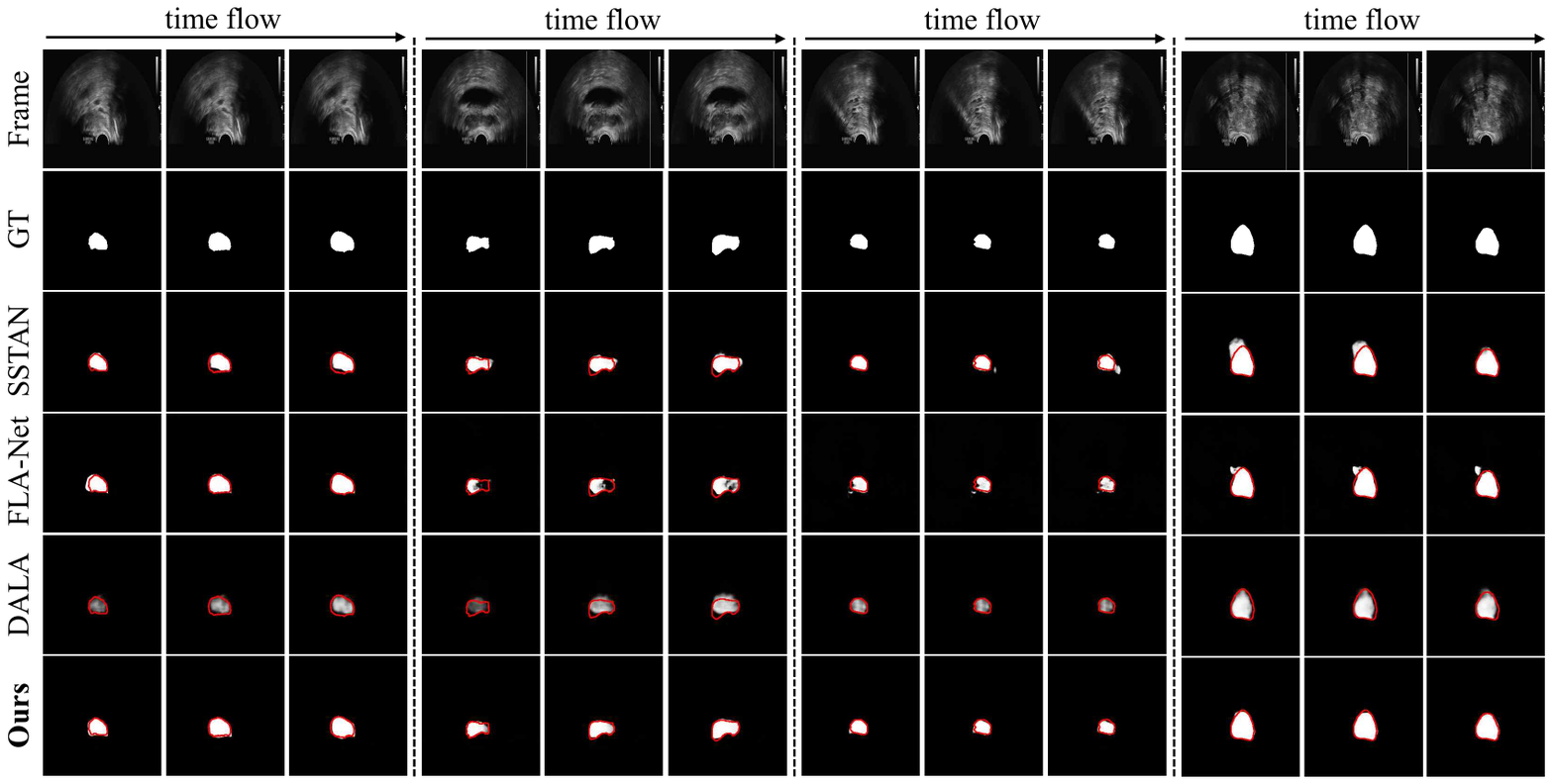}
    \vspace{-6mm}
    \caption{\textbf{Visual comparison on TRUS-V.} Our method maintains robust boundaries aligned with the ground truth (red contours) across consecutive frames even under acoustic shadows, whereas competitors suffer from under-segmentation.}
    \label{fig:5}
    \vspace{-3mm}
\end{figure}

\noindent\textbf{Performance on Clinical TRUS-V.} We further validate our method on the in-house TRUS-V benchmark (Table~\ref{tab:trus_v}). Interestingly, we observed that existing video segmentation models (SSTAN~\cite{zhao2022semi}, FLA-Net~\cite{lin2023shifting}, DALA~\cite{zhao2025efficient}), which generally excel in colonoscopy video scenes, struggle to outperform strong 2D baselines in the ultrasound domain. This performance degradation is likely attributed to the severe speckle noise and ambiguous boundaries inherent in TRUS, which disrupt standard temporal attention mechanisms. In contrast, our framework effectively handles these domain-specific challenges, achieving state-of-the-art performance across metrics at a real-time 127.97 FPS (U-Net$^{++}$~\cite{zhou2019unet++}). These quantitative gains are corroborated by the visual comparisons in Fig.~\ref{fig:5}; while competing methods often produce intermittent false negatives under acoustic shadows or rapid probe motion, our model maintains smooth and anatomically plausible boundaries. This confirms that our temporal coherence distillation successfully learns robust representations that are resilient to signal dropouts.

\subsection{Ablation Studies}


As shown in Table~\ref{tab:ablation}, $\mathcal{L}_{KD}$ stabilizes the noisy
baseline (Dice 0.171$\to$0.701) but lacks temporal coherence. Individually, the
pixel-level $\mathcal{L}_{con}$ and the dual-scale $\mathcal{L}_{proto}$ raise Dice
to 0.722 and 0.735, respectively. Here, $\triangle$ denotes single-scale usage:
local-only in ``w/o Global Proto.'' and global-only in ``w/o Local Proto.'' These
fail to improve over $\mathcal{L}_{con}$ (0.721/0.722 vs.\ 0.722), as local
prototypes are noise-sensitive while global ones are too coarse for boundaries.
Their area-adaptive combination yields the best performance ($S_{\alpha}$ 0.816,
Dice 0.746).

\section{Conclusion}

In this paper, we presented a novel Temporally Consistent Learning Framework for robust prostate video segmentation. By successfully distilling temporal coherence through motion-aware consistency and prototype alignment, our approach effectively transfers temporal knowledge to a lightweight 2D backbone, solving the flickering problem inherent in slice-based models. Crucially, our self-supervised consistency strategy, empowered by a Student-Teacher architecture, enables the effective utilization of unlabeled video data while mitigating catastrophic forgetting. We also released TRUS-V, a comprehensive multi-view benchmark, to facilitate future research in this domain. Extensive experiments on both TRUS-V and SUN-SEG datasets demonstrated that our method achieves state-of-the-art performance, offering a promising solution for real-time, temporally stable image-guided prostate intervention. Future work will focus on extending this framework to 3D volume reconstruction and real-time biopsy guidance systems.

\subsubsection*{Acknowledgements.}
This work was supported by the National Research Foundation of Korea (NRF) grant funded by the Korea government (MSIT) (Grant No. RS-2025-00553835 and RS-2024-00343630), and a grant of the Korea Health Technology R\&D Project through the Korea Health Industry Development Institute (KHIDI), funded by the Ministry of Health \& Welfare, Republic of Korea (Grant No. RS-2025-02307233).

\noindent\textbf{Disclosure of Interests.}
The authors have no competing interests to declare that are relevant to the content of this article.
%
%
%
\bibliographystyle{splncs04}
\bibliography{cas-refs}

\end{document}